\documentclass[letterpaper]{article} % DO NOT CHANGE THIS
\usepackage{aaai2026}  % DO NOT CHANGE THIS
\usepackage{times}  % DO NOT CHANGE THIS
\usepackage{helvet}  % DO NOT CHANGE THIS
\usepackage{courier}  % DO NOT CHANGE THIS
\usepackage[hyphens]{url}  % DO NOT CHANGE THIS
\usepackage{graphicx} % DO NOT CHANGE THIS
\usepackage{natbib}  % DO NOT CHANGE THIS AND DO NOT ADD ANY OPTIONS TO IT
\usepackage{caption} % DO NOT CHANGE THIS AND DO NOT ADD ANY OPTIONS TO IT
\usepackage{algorithm}
\usepackage{algorithmic}

\usepackage{amsmath, amssymb}
\usepackage{multirow}
\usepackage{booktabs}
\usepackage{graphicx}
\usepackage{caption}
\usepackage{adjustbox}
\usepackage{pifont}

\newcommand{\cmark}{\ding{51}}  % √
\newcommand{\xmark}{\ding{55}}  % ✗

\usepackage{newfloat}
\usepackage{listings}
\DeclareCaptionStyle{ruled}{labelfont=normalfont,labelsep=colon,strut=off} % DO NOT CHANGE THIS
\floatstyle{ruled}
\newfloat{listing}{tb}{lst}{}
\floatname{listing}{Listing}
\nocopyright
\title{NeuroBridge: Bio-Inspired Self-Supervised EEG-to-Image Decoding\\via Cognitive Priors and Bidirectional Semantic Alignment}
\author {
    Wenjiang Zhang\textsuperscript{\rm 1},
    Sifeng Wang\textsuperscript{\rm 1},
    Yuwei Su\textsuperscript{\rm 1},
    Xinyu Li\textsuperscript{\rm 1},
    Chen Zhang\textsuperscript{\rm 1},
    Suyu Zhong\textsuperscript{\rm 1, 2}\footnote{Corresponding author.}
}
\affiliations {
    \textsuperscript{\rm 1}School of Artificial Intelligence, Beijing University of Posts and Telecommunications, Beijing, China\\
    \textsuperscript{\rm 2}Queen Mary School Hainan, Beijing University of Posts and Telecommunications, Hainan 570100, China\\
    wjzhang@bupt.edu.cn,
    zhongsuyu@bupt.edu.cn
}
\begin{document}

\maketitle

\begin{abstract}
Visual neural decoding seeks to reconstruct or infer perceived visual stimuli from brain activity patterns, providing critical insights into human cognition and enabling transformative applications in brain-computer interfaces and artificial intelligence. Current approaches, however, remain constrained by the scarcity of high-quality stimulus-brain response pairs and the inherent semantic mismatch between neural representations and visual content. Inspired by perceptual variability and co-adaptive strategy of the biological systems, we propose a novel self-supervised architecture, named NeuroBridge, which integrates Cognitive Prior Augmentation (CPA) with Shared Semantic Projector (SSP) to promote effective cross-modality alignment. Specifically, CPA simulates perceptual variability by applying asymmetric, modality-specific transformations to both EEG signals and images, enhancing semantic diversity. Unlike previous approaches, SSP establishes a bidirectional alignment process through a co-adaptive strategy, which mutually aligns features from two modalities into a shared semantic space for effective cross-modal learning. NeuroBridge surpasses previous state-of-the-art methods under both intra-subject and inter-subject settings. In the intra-subject scenario, it achieves the improvements of \textbf{12.3\%} in Top-1 accuracy and \textbf{10.2\%} in Top-5 accuracy, reaching \textbf{63.2\%} and \textbf{89.9\%} respectively on a 200-way zero-shot retrieval task. Extensive experiments demonstrate the effectiveness, robustness, and scalability of the proposed framework for neural visual decoding.
\end{abstract}

\begin{links}
    \link{Code}{https://github.com/feroooooo/NeuroBridge}
\end{links}

\section{Introduction}
Understanding how the human brain processes visual information is a fundamental challenge in both cognitive neuroscience and artificial intelligence~\cite{posner_attention_1980, kamitani_decoding_2005, gao_interface_2021}. Visual neural decoding, which interprets neural signals evoked by visual stimuli (e.g., images) to infer perceptual and cognitive states, holds significant promise for unraveling the mechanisms of visual processing and advancing brain-inspired AI. While a variety of non-invasive neuroimaging techniques, such as functional magnetic resonance imaging (fMRI) ~\cite{mindeye, mindeye2, chen2024bridging, chen2024mind, dai2025mindaligner} and magnetoencephalography (MEG)~\cite{benchetrit2024brain} enable the capture of dynamic brain activity in response to visual inputs, electroencephalography (EEG) stands out due to its excellent temporal resolution, high cost-effectiveness and high portability~\cite{du2023decoding}. 

Cross-modal contrastive learning between EEG signals and visual images has emerged as the predominant approach for EEG-based visual neural decoding by establishing semantic alignment between modalities~\cite{du2023decoding, song2024decoding, wu2025bridging}. The effectiveness of cross-modal learning is predominantly governed by the degree of achieved semantic alignment between modalities. However, an inherent modality gap exists between EEG signals and images, which manifests in two fundamental levels: dynamic variability gap and static intrinsic gap. More specifically, the dynamic gap encompasses intra- and inter-subject EEG  variability to identical images, attributable to probabilistic neuro-cognitive factors (e.g., attentional fluctuations, mental states and physiological noise) alongside inherent biological noise~\cite{boynton2005attention}. In short, this dynamic fluctuation reflects inter- and intra-individual perceptual variability inherent to EEG neural dynamics. The static intrinsic EEG-image gap originates from the irreducible divergence between temporal, low-dimensional, noise-prone EEG signals and spatially structured, high-dimensional, semantically dense images~\cite{piastra_comprehensive_2021}. These intrinsic mismatches hinder effective cross-modal alignment. Although EEG datasets have grown in size~\cite{grootswagers_human_2022, gifford2022large}, they remain limited compared to the large, diverse datasets in computer vision~\cite{imagenet}, further widening the gap and necessitating stronger architectural priors for robust, generalizable decoding.

\begin{figure}[htbp]
\centering
\includegraphics[width=\linewidth]{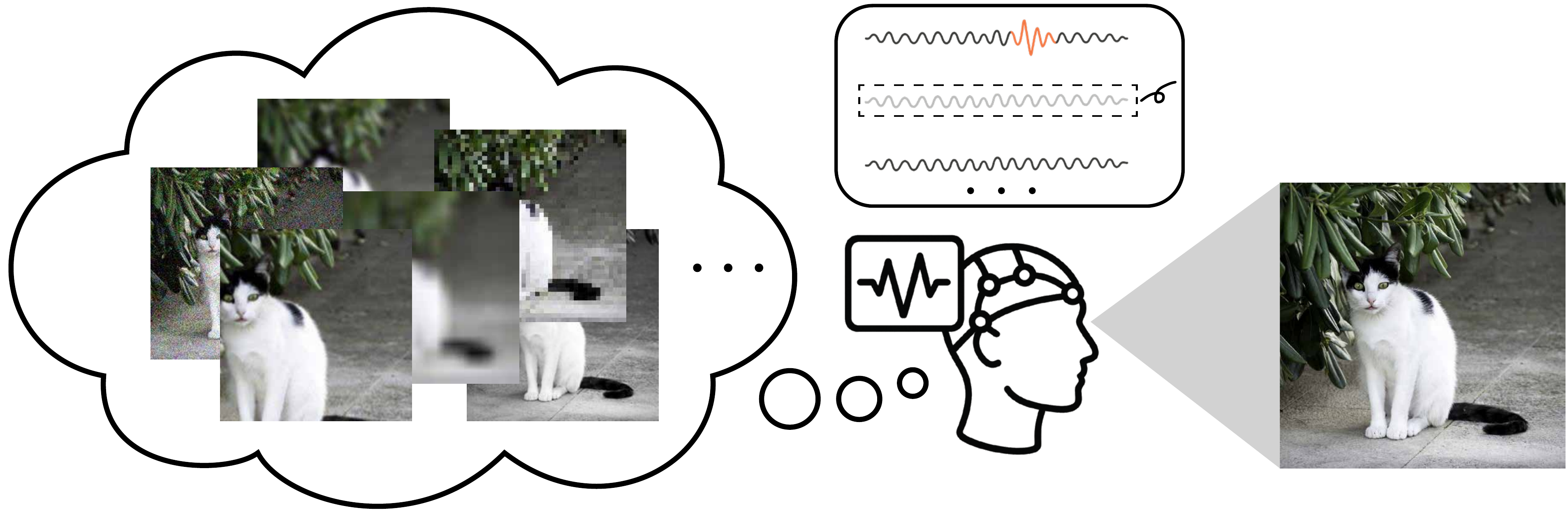}
\caption{Illustration of human visual cognition and EEG variability. When viewing a visual stimulus, individuals focus on different semantic regions (e.g., parts of a cat), leading to varied EEG responses influenced by attention, mental state, and individual differences.}
\label{fig:cognition}
\end{figure}

To address this challenge, we propose a novel framework, called NeuroBridge, inspired by perceptual variability and co-adaptive strategy of the biological systems, to bridge EEG-based neural and visual modalities. NeuroBridge integrates Cognitive Prior Augmentation (CPA) and Shared Semantic Projector (SSP). CPA enriches the learning process by simulating the human cognitive response during visual perception. As mentioned above,  EEG signals are characterized by perceptual-state-dependent properties (Fig.~\ref{fig:cognition}). By leveraging the perceptual variability of the EEG, CPA transforms both EEG signals and images. Besides, considering the static inherent gap between EEG and image modalities (i.e. data structure, feature scale, and semantic richness), CPA also employs an asymmetric augmentation strategy tailored to each modality. This enables more targeted enhancement and alignment, helping to bridge the modality gap more effectively. Considering that conventional EEG-based visual decoding employs unidirectional alignment, which prioritizes computational efficiency yet is prone to semantic incongruence, we propose SSP with the biologically inspired co-adaptation mechanism\ \cite{song2024decoding}. Specifically, SSP complements cross-modal semantic alignment by projecting modality-specific features into a unified semantic space. This shared semantic projection not only aligns multimodal features but also enhances cross-modal interaction. While the image encoder is typically pretrained and frozen, the optimization of SSP allows EEG features to be more effectively aligned with the rich semantic representations derived from large-scale visual datasets.

Overall, NeuroBridge provides a principled and efficient approach to bridging the modality gap between EEG and visual data, demonstrating strong decoding performance and good scalability. Our main contributions are summarized as follows:

\begin{enumerate}
\item We present \textbf{NeuroBridge}, a unified and efficient framework that achieves strong performance and robust generalization across diverse settings by effectively bridging the modality gap between EEG signals and visual representations.
\item We propose \textbf{Cognitive Prior Augmentation (CPA)}, a powerful strategy that simulates cognitive variations, enhancing the diversity of both EEG and image features.
\item We design a \textbf{Shared Semantic Projector (SSP)} that efficiently maps multimodal features into a common semantic space, enabling more effective cross-modal learning.
\item Extensive experiments demonstrate the strong performance and generalization of our method. On the zero-shot brain-to-image retrieval task, our framework achieves \textbf{63.2\%} Top-1 and \textbf{89.9\%} Top-5 accuracy, surpassing previous state-of-the-art results by large margins (\textbf{+12.3\%} and \textbf{+10.2\%}, respectively).
\end{enumerate}

\section{Related Work}
\subsection{Multimodal Contrastive Representation Learning}
Multimodal contrastive learning focuses on aligning representations from different modalities, such as vision and language, within a shared embedding space.

Multimodal contrastive representation learning is fundamentally predicated on cross-modal alignment, the process of projecting heterogeneous modalities (e.g., vision, language) into a shared embedding space where semantically congruent instances are jointly optimized through contrastive objectives. In recent years, numerous multimodal contrastive learning approaches have achieved impressive results on large-scale datasets~\cite{schuhmann2021laion}. Representative works such as CLIP~\cite{Radford2021LearningTV}, ALIGN~\cite{pmlr-v139-jia21b}, and BLIP~\cite{pmlr-v162-li22n} leverage hundreds of millions of image-text pairs to train dual-encoder architectures that align modalities in a shared embedding space. These models exhibit strong zero-shot performance across a wide range of downstream tasks, including image-text retrieval, classification, and visual question answering, demonstrating the effectiveness of large-scale contrastive pretraining for learning generalizable multimodal representations when sufficient training pairs exist. Notably, this success hinges critically on the (1) sufficient quantities of paired samples to cover the semantic space, and (2) rigorous semantic alignment that preserves conceptual relationships across modalities.

However, certain modality pairs still face significant challenges. For instance, audio-image~\cite{audioclip}, video-text~\cite{Bain_2021_ICCV}, 3D-language~\cite{Xue_2023_CVPR}, and EEG-image all lack large-scale, high-quality aligned datasets. These modalities often exhibit greater heterogeneity or weaker natural alignment. EEG-image, in particular, suffers from the complexity of neural signals, which hinders effective cross-modal correspondence. As a result, models trained in these settings often struggle to generalize, especially in zero-shot scenarios. Addressing data scarcity and improving alignment quality remain key open problems in contrastive multimodal representation learning.

\begin{figure*}[t]
    \centering
    \includegraphics[width=\textwidth]{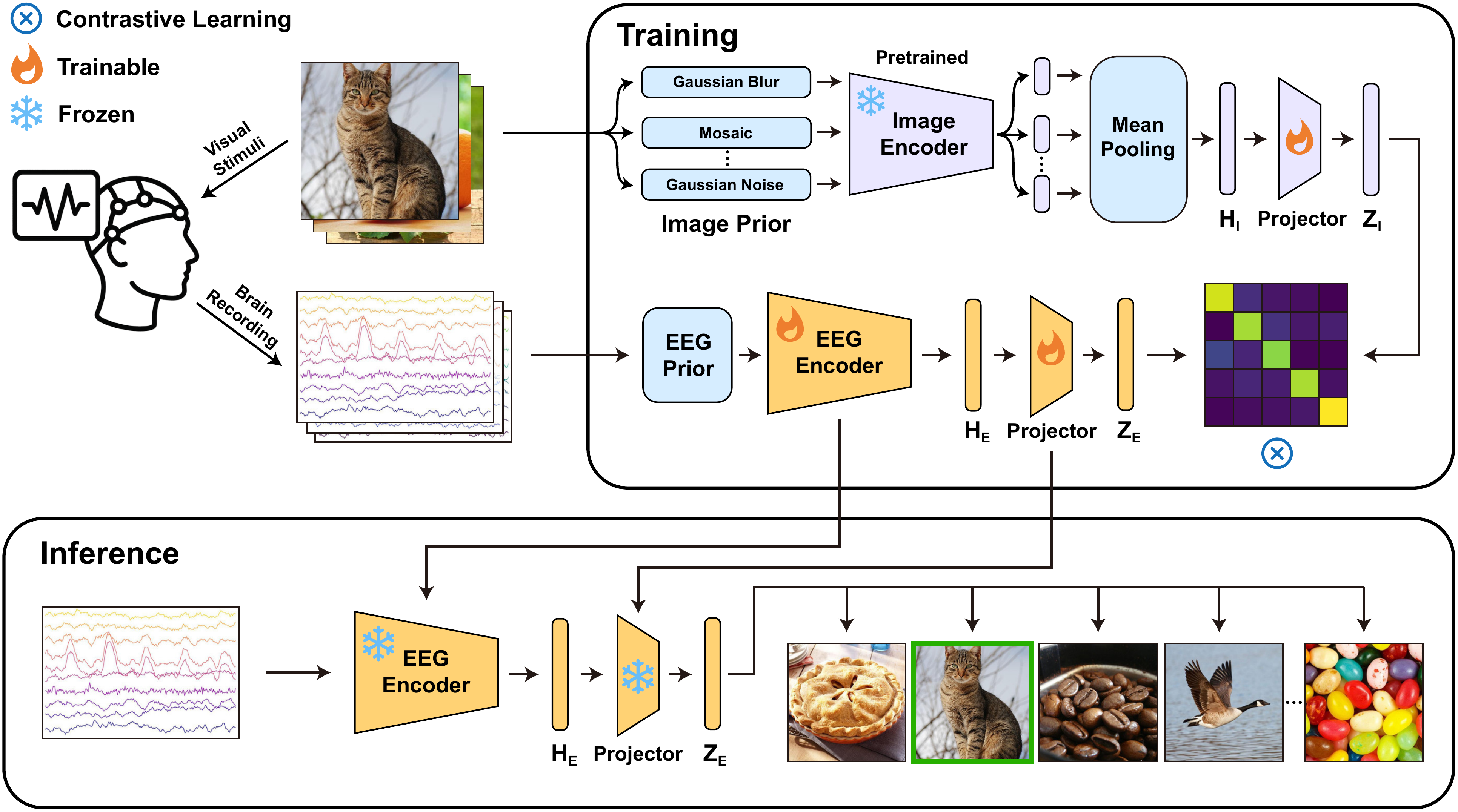}
    \caption{Overall framework of NeuroBridge. During training, cognitively augmented image-EEG pairs are encoded and projected into a shared semantic space using contrastive learning. At inference, EEG signals are decoded by matching their embeddings to visual concepts.}
    \label{fig:overall}
\end{figure*}

\subsection{Neural Visual Decoding}
Neural visual decoding aims to reconstruct or interpret visual content perceived by humans from brain activity, such as fMRI, MEG or EEG signals. EEG has become pivotal in neural decoding for capturing dynamic visual processing, owing to its millisecond temporal resolution. In the realm of EEG-based decoding, one pioneering work~\cite{Spampinato_2017_CVPR} demonstrated significant potential for visual recognition tasks, sparking a wave of subsequent research. However, the block-design paradigm used in their dataset, which grouped images of the same class together, has been criticized for enabling models to exploit low-level statistical cues instead of genuine neural representations~\cite{perils, Ahmed_2021_CVPR}. Consequently, the performance of methods based on the dataset using the block-design paradigm remains subject to further scrutiny~\cite{kavasidis2017brain2image, thoughtviz, dreamdiffusion}. In contrast, the Rapid Serial Visual Presentation (RSVP) paradigm mitigates these issues by presenting stimuli in a randomized and fast-paced manner, reducing experimental bias~\cite{GROOTSWAGERS2019668}. Large-scale datasets based on RSVP, such as THINGS-EEG~\cite{grootswagers_human_2022} and THINGS-EEG2~\cite{gifford2022large}, have since become essential benchmarks for EEG-based neural visual decoding research~\cite{du2023decoding, song2024decoding, wu2025bridging}.

For the model, although many studies have explored joint modeling of image and EEG modalities, few have explicitly addressed the semantic and representational discrepancies between them. Besides, UBP~\cite{wu2025bridging} introduced a blur prior to capture early-stage perceptual information in the visual modality, while Neural-MCRL~\cite{li2024neural} proposed intra-modal semantic completion to enhance semantic representation within the EEG modality. Although these methods partially reflect underlying cognitive processes, they predominantly focus on unimodal enhancements such as blur priors for visual refinement and semantic completion for EEG, lacking a comprehensive bimodal framework for synergistic cross-modal modeling.

\section{Method}
We propose \textbf{NeuroBridge}, a simple yet effective self-supervised EEG-to-image decoding framework with strong extensibility. It consists of two key modules, that is CPA and SSP. The overall architecture is illustrated in Figure~\ref{fig:overall}.

During training, paired EEG-image data are used to extract modality-specific representations. A contrastive learning objective aligns matched pairs while pushing apart mismatched ones. After training, the encoders and projectors enable efficient inference. The full algorithmic flow is outlined in Algorithm~\ref{alg:framework}.

\begin{algorithm}[t]
  \caption{Training Algorithm of NeuroBridge}
  \label{alg:framework}

  {
  \setlength{\baselineskip}{1.28\baselineskip}

  \begin{algorithmic}[1]
    \REQUIRE Paired data $D_{\text{train}}=\{(x_I,x_E)\}$; image transformations $\{t_{I,k}\}_{k=1}^{K}$; EEG transformation $t_E$; encoders $f_I$, $f_E$; projectors $p_I$, $p_E$; temperature $\tau$; optimizer $\mathcal{O}$
    \ENSURE Trained $f_E$, $p_I$, $p_E$
    \FOR{\textbf{each} batch $(X_I, X_E) \in D_{\text{train}}$}
        \STATE $X_{I',k}=t_{I,k}(X_I)$
        \STATE $X_{E'}=t_E(X_E)$
        \STATE $H_{I,k}=f_I(X_{I',k})$
        \STATE $H_E=f_E(X_{E'})$
        \STATE $H_I=\frac{1}{K}\sum_{k=1}^{K} H_{I,k}$
        \STATE $Z_I=\operatorname{norm}(p_I(H_I))$, $Z_E=p_E(H_E)$
        \STATE $L=\mathrm{ContrastiveLoss}(Z_I,Z_E;\tau)$
        \STATE Update $\theta_{f_E},\theta_{p_I},\theta_{p_E}$ with $\mathcal{O}(\nabla_\theta L)$
    \ENDFOR
  \end{algorithmic}

  }
\end{algorithm}

\subsection{Problem Definition}
Our task is zero-shot neural visual decoding, following the setup established in prior work~\cite{song2024decoding}. Let the image data be denoted as $x_I \in \mathbb{R}^{C_I \times H \times W}$ and the corresponding EEG data as $x_E \in \mathbb{R}^{C_E \times T}$, where $C_I$ is the number of image channels, $H$ and $W$ are image height and width, $C_E$ is the number of EEG channels, and $T$ denotes the time window length. Denote the complete sets of image and EEG data as $X_I$ and $X_E$, respectively. The paired dataset is represented as $D = \{(x_I^{(n)}, x_E^{(n)})\}_{n=1}^{N}$, which is split into training and test sets $D_{\text{train}}$ and $D_{\text{test}}$, with $D_{\text{train}} \cap D_{\text{test}} = \varnothing$.

During training, the model is optimized using $D_{\text{train}}$. In the testing phase, $X_I^{test} = \{x_I^{(m)}\}_{m=1}^{M}$ is used as the visual concept pool, and EEG features from $X_E^{test} = \{x_E^{(m)}\}_{m=1}^{M}$ are decoded by computing their similarities to the candidate visual features.

\subsection{Cognitive Prior Augmentation}
We posit that many augmentations or transformations of images and EEG signals can serve as cognitive priors, providing auxiliary cues during visual interpretation. To this end, we introduce CPA, a modality-specific augmentation strategy designed to incorporate cognitively relevant priors into contrastive learning. Unlike conventional augmentations primarily designed for classification tasks, CPA aims to simulate human perceptual invariance to transformations in both visual and neural domains.

Given the rich prior knowledge embedded in pretrained image encoders, multiple augmentation strategies are applied to image data. Conversely, due to the absence of such priors in EEG data, an asymmetric design is adopted, where only a single augmentation is applied to EEG signals.

The augmentation process is defined as:

\begin{equation}
    X_{I', k} = t_{I, k}(X_I),\quad X_{E'} = t_E(X_E)
\end{equation}

Here, $t_{I, k}$ and $t_E$ denote image and EEG transformations, respectively, where $k$ indexes different image transformation strategies. CPA offers a flexible augmentation strategy, where specific transformations can be tailored to the requirements of different tasks. The specific augmentation methods are introduced in the following sections.

\subsection{Image and EEG Feature Extraction}
We employ a frozen pretrained image encoder (i.e., CLIP) to extract visual features, while the EEG encoder is trained from scratch. The architecture of both encoders is modular and can be customized.

Due to the large-scale architecture and rich semantic knowledge of the pretrained image encoder, leveraging multiple augmented views is essential to fully exploit its capabilities.

The encoding process is defined as:

\begin{equation}
    H_{I, k} = f_I(X_{I', k}),\quad H_E = f_E(X_{E'})
\end{equation}

where $f_I$ and $f_E$ denote the image and EEG encoders, and $H_{I,k}$, $H_E$ are the corresponding features.

\subsection{Semantic Aggregation of Image Representations}
Each image transformation produces a distinct feature. Instead of using all features independently, which would significantly increase computational cost and complicate feature alignment, we perform feature fusion to obtain a single representative embedding.

Given that the image encoder is frozen and pretrained, its output features reside in a stable semantic space. Therefore, we average the features obtained from multiple augmented views to generate a compact and semantically enriched representation:

\begin{equation}
    H_I = \frac{1}{K}\sum_{k=1}^KH_{I,k}
\end{equation}

where $K$ denotes the number of applied image augmentation strategies.

\subsection{Shared Semantic Projector}
Pretrained vision-language models, such as CLIP, learn semantic spaces shaped by large-scale image-text associations and are heavily influenced by linguistic co-occurrence. In contrast, EEG signals reflect perceptual and cognitive processes rooted in neural dynamics, which differ fundamentally from language-driven semantics. This inherent modality gap presents significant challenges for direct alignment between EEG features and pretrained visual representations.

To address this issue, we introduce the SSP, which maps both image and EEG features into a unified, trainable semantic space. This modality-agnostic space enables effective alignment by learning semantic correspondence from data, rather than relying on a fixed prior. The projection networks within SSP can be implemented as either linear or nonlinear transformations, offering flexibility to adapt to different tasks or datasets.

The projection process is formulated as:

\begin{equation}
    Z_I = p_I(H_I),\quad Z_E = p_E(H_E)
\end{equation}

where $p_I$ and $p_E$ are the projection networks for the image and EEG modalities, respectively. $Z_I$ and $Z_E$ denote their corresponding embeddings in the shared semantic space.

\subsection{Modality-Aware Contrastive Learning}
We adopt a modality-aware contrastive learning method to align image and EEG features by bringing matched pairs closer and pushing apart mismatched pairs. The loss is defined as:

\begin{align}
    L = \frac{1}{2N}\sum_{j=1}^N \bigg[
        & -\log\frac{\exp(sim(z_{I,j}, z_{E,j}) / \tau)}{\sum_{k=1}^N \exp(sim(z_{I,j}, z_{E,k}) / \tau)} \notag \\
        & -\log\frac{\exp(sim(z_{E,j}, z_{I,j}) / \tau)}{\sum_{k=1}^N \exp(sim(z_{E,j}, z_{I,k}) / \tau)}
    \bigg]
\end{align}

\begin{table*}[t]
\centering
\small
\begin{tabular}{lc|cccccccccc|c}
\toprule
\multicolumn{2}{l|}{Method} & Sub 1 & Sub 2 & Sub 3 & Sub 4 & Sub 5 & Sub 6 & Sub 7 & Sub 8 & Sub 9 & Sub 10 & Average \\
\midrule
\multicolumn{13}{c}{\textbf{Intra-Subject: train and test on one subject}} \\
\midrule
\multirow{2}{*}{BraVL} & Top-1 & 6.1 & 4.9 & 5.6 & 5.0 & 4.0 & 6.0 & 6.5 & 8.8 & 4.3 & 7.0 & 5.8\\
& Top-5 & 17.9 & 14.9 & 17.4 & 15.1 & 13.4 & 18.2 & 20.4 & 23.7 & 14.0 & 19.7 & 17.5\\
\midrule
\multirow{2}{*}{NICE} & Top-1 & 13.2 & 13.5 & 14.5 & 20.6 & 10.1 & 16.5 & 17.0 & 22.9 & 15.4 & 17.4 & 16.1\\
& Top-5 & 39.5 & 40.3 & 42.7 & 52.7 & 31.5 & 44.0 & 42.1 & 56.1 & 41.6 & 45.8 & 43.6\\
\midrule
\multirow{2}{*}{ATM} & Top-1 & 25.6 & 22.0 & 25.0 & 31.4 & 12.9 & 21.3 & 30.5 & 38.8 & 34.4 & 29.1 & 27.1\\
& Top-5 & 60.4 & 54.5 & 62.4 & 60.9 & 43.0 & 51.1 & 61.5 & 72.0 & 51.5 & 63.5 & 58.1\\
\midrule
\multirow{2}{*}{CognitionCapturer} & Top-1 & 27.2 & 28.7 & 37.2 & 37.7 & 21.8 & 31.6 & 32.8 & 47.6 & 33.4 & 35.1 & 33.3\\
& Top-5 & 59.5 & 57.0 & 66.1 & 63.2 & 47.8 & 58.1 & 59.6 & 73.5 & 57.7 & 63.6 & 60.6\\
\midrule
\multirow{2}{*}{Neural-MCRL} & Top-1 & 27.5 & 28.5 & 37.0 & 35.0 & 22.5 & 31.5 & 31.5 & 42.0 & 30.5 & 37.5 & 32.4\\
& Top-5 & 64.0 & 61.5 & 69.0 & 66.0 & 51.5 & 61.0 & 62.5 & 74.5 & 59.5 & 71.0 & 64.1\\
\midrule
\multirow{2}{*}{VE-SDN} & Top-1 & 32.6 & 34.4 & 38.7 & 39.8 & 29.4 & 34.5 & 34.5 & 49.3 & 39.0 & 39.8 & 37.2\\
& Top-5 & 63.7 & 69.9 & 73.5 & 72.0 & 58.6 & 68.8 & 68.3 & 79.8 & 69.6 & 75.3 & 70.0\\
\midrule
\multirow{2}{*}{UBP} & Top-1 & 41.2 & 51.2 & 51.2 & 51.1 & 42.2 & 57.5 & 49.0 & 58.6 & 45.1 & 61.5 & 50.9\\
& Top-5 & 70.5 & 80.9 & 82.0 & 76.9 & 72.8 & 83.5 & 79.9 & 85.8 & 76.2 & 88.2 & 79.7\\
\midrule
\multirow{2}{*}{\textbf{NeuroBridge (Ours)}} & Top-1 & \textbf{50.0} & \textbf{63.2} & \textbf{61.6} & \textbf{61.4} & \textbf{54.8} & \textbf{69.7} & \textbf{62.7} & \textbf{71.2} & \textbf{64.0} & \textbf{73.6} & \textbf{63.2}\\
& Top-5 & \textbf{77.6} & \textbf{90.6} & \textbf{91.1} & \textbf{90.0} & \textbf{85.0} & \textbf{92.9} & \textbf{88.8} & \textbf{95.1} & \textbf{91.0} & \textbf{97.1} & \textbf{89.9}\\
\midrule
\multicolumn{13}{c}{\textbf{Inter-Subject: leave one subject out for test}} \\
\midrule
\multirow{2}{*}{BraVL} & Top-1 & 2.3 & 1.5 & 1.4 & 1.7 & 1.5 & 1.8 & 2.1 & 2.2 & 1.6 & 2.3 & 1.8\\
& Top-5 & 8.0 & 6.3 & 5.9 & 6.7 & 5.6 & 7.2 & 8.1 & 7.6 & 6.4 & 8.5 & 7.0\\
\midrule
\multirow{2}{*}{NICE} & Top-1 & 7.6 & 5.9 & 6.0 & 6.3 & 4.4 & 5.6 & 5.6 & 6.3 & 5.7 & 8.4 & 6.2\\
& Top-5 & 22.8 & 20.5 & 22.3 & 20.7 & 18.3 & 22.2 & 19.7 & 22.0 & 17.6 & 28.3 & 21.4\\
\midrule
\multirow{2}{*}{ATM} & Top-1 & 10.5 & 7.1 & 11.9 & 14.7 & 7.0 & 11.1 & \textbf{16.1} & 15.0 & 4.9 & 20.5 & 11.9\\
& Top-5 & 26.8 & 24.8 & 33.8 & 39.4 & 23.9 & 35.8 & 43.5 & 40.3 & 22.7 & 46.5 & 33.8\\
\midrule
\multirow{2}{*}{UBP} & Top-1 & 11.5 & 15.5 & 9.8 & 13.0 & 8.8 & 11.7 & 10.2 & 12.2 & \textbf{15.5} & 16.0 & 12.4\\
& Top-5 & 29.7 & 40.0 & 27.0 & 32.3 & 33.8 & 31.0 & 23.8 & 32.2 & \textbf{40.5} & 43.5 & 33.4\\
\midrule
\multirow{2}{*}{Neural-MCRL} & Top-1 & 13.0 & 12.0 & \textbf{14.5} & 12.5 & 11.5 & 13.5 & 14.0 & 18.5 & 13.5 & 17.0 & 14.0\\
& Top-5 & 31.5 & 30.5 & 35.5 & 35.5 & 29.0 & 35.5 & 36.0 & 38.5 & 32.5 & 39.0 & 34.3\\
\midrule
\multirow{2}{*}{\textbf{NeuroBridge (Ours)}} & Top-1 & \textbf{23.2} & \textbf{21.2} & 13.2 & \textbf{17.0} & \textbf{14.5} & \textbf{25.0} & 15.3 & \textbf{20.1} & 13.7 & \textbf{27.2} & \textbf{19.0}\\
& Top-5 & \textbf{52.4} & \textbf{49.3} & \textbf{36.5} & \textbf{45.3} & \textbf{37.7} & \textbf{55.0} & \textbf{45.1} & \textbf{44.9} & 36.5 & \textbf{56.3} & \textbf{45.9}\\
\bottomrule
\end{tabular}
\caption{Overall accuracy (\%) of 200-way zero-shot retrieval on THINGS-EEG2: Top-1 and Top-5.}
\label{table:main}
\end{table*}

Here, $z_{I,j} \in Z_I$ and $z_{E,j} \in Z_E$ are paired image and EEG embeddings; $N$ is the number of paired samples in the current mini-batch; and $\tau$ is the temperature. The similarity function $sim(\cdot,\cdot)$ uses a modified cosine similarity that omits EEG normalization.

Unlike symmetric contrastive learning, our method normalizes image features to the unit hypersphere while allowing EEG feature magnitudes to vary. This leverages feature direction for semantic alignment and magnitude for learnable confidence, preserving the pretrained image structure while improving EEG robustness and training stability.

\section{Experiments and Results}

\subsection{Datasets and Preprocessing}
We conducted our experiments on the THINGS-EEG2 dataset~\cite{gifford2022large}, which contains recordings from 10 subjects using the RSVP paradigm. The training set comprises 1,654 concepts, each associated with 10 images, with each image repeated four times per subject. The test set consists of 200 concepts, each represented by one image, with each image repeated 80 times per subject.

For data preprocessing, we follow the methodology described in previous work~\cite{song2024decoding, li2024visual, wu2025bridging}. EEG data are segmented into 0-1,000 ms trials following stimulus onset, with baseline correction using the average signal from the 200 ms preceding the stimulus. The data are then downsampled to 250 Hz. Multivariate noise normalization (MVNN) is applied for normalization. Repetitions are averaged to enhance SNR, resulting in a total of 16,540 training samples and 200 test samples per subject.

See the Appendix for additional evaluation on THINGS-MEG~\cite{hebart2023things}.

\subsection{Implementation Details}
Our method is implemented using PyTorch and trained on two NVIDIA GeForce RTX 3090 GPUs. We employ a batch size of 1,024 and train the model for 50 epochs. The learning rate is set to 1e-4. Gradient updates are performed using the AdamW optimizer with a weight decay of 1e-4. The temperature parameter $\tau$ is set to 0.07. All reported results represent the average over five independent training runs.

\subsubsection{Encoder}
For the image encoder, we adopt CLIP~\cite{Radford2021LearningTV} with weights from OpenCLIP~\cite{cherti2023reproducible}, utilizing various backbones including RN50, RN101, ViT-B-16, ViT-B-32, ViT-L-14, ViT-H-14, ViT-g-14, and ViT-bigG-14, with RN50 as default. For the brain encoder, we implement EEGNet~\cite{lawhern_eegnet_2018}, TSConv~\cite{song2024decoding}, ATM~\cite{li2024visual}, and EEGProject~\cite{wu2025bridging}, where EEGProject is the default.

\subsubsection{CPA}
For image augmentation, we employ Gaussian blur, Gaussian noise, low resolution, mosaic, color jitter, grayscale, and random crop. EEG augmentation includes channel dropout, noise addition, smoothing, and temporal shifting. By default, Gaussian blur, Gaussian noise, low resolution and mosaic are used for image prior, while smoothing is used for EEG prior.

\subsubsection{SSP}
We experimented with both linear projection and MLPs across various feature dimensions.

\subsection{Comparison with Baselines}
We compared our method with several recent works: BraVL~\cite{du2023decoding}, NICE~\cite{song2024decoding}, ATM~\cite{li2024visual}, CognitionCapturer~\cite{zhang2025cognitioncapturer}, Neural-MCRL~\cite{li2024neural}, VE-SDN~\cite{chen2024visual}, and UBP~\cite{wu2025bridging}. Table~\ref{table:main} presents the quantitative results on the THINGS-EEG2 dataset. Our method surpasses the state-of-the-art in both intra-subject and inter-subject evaluations. Specifically, our approach achieves a Top-1 accuracy of 63.2\% and a Top-5 accuracy of 89.9\% in the zero-shot brain-image retrieval task.

\subsection{Effect of Image and EEG Transformations}
\begin{figure}[htbp]
    \centering
    \includegraphics[width=\linewidth]{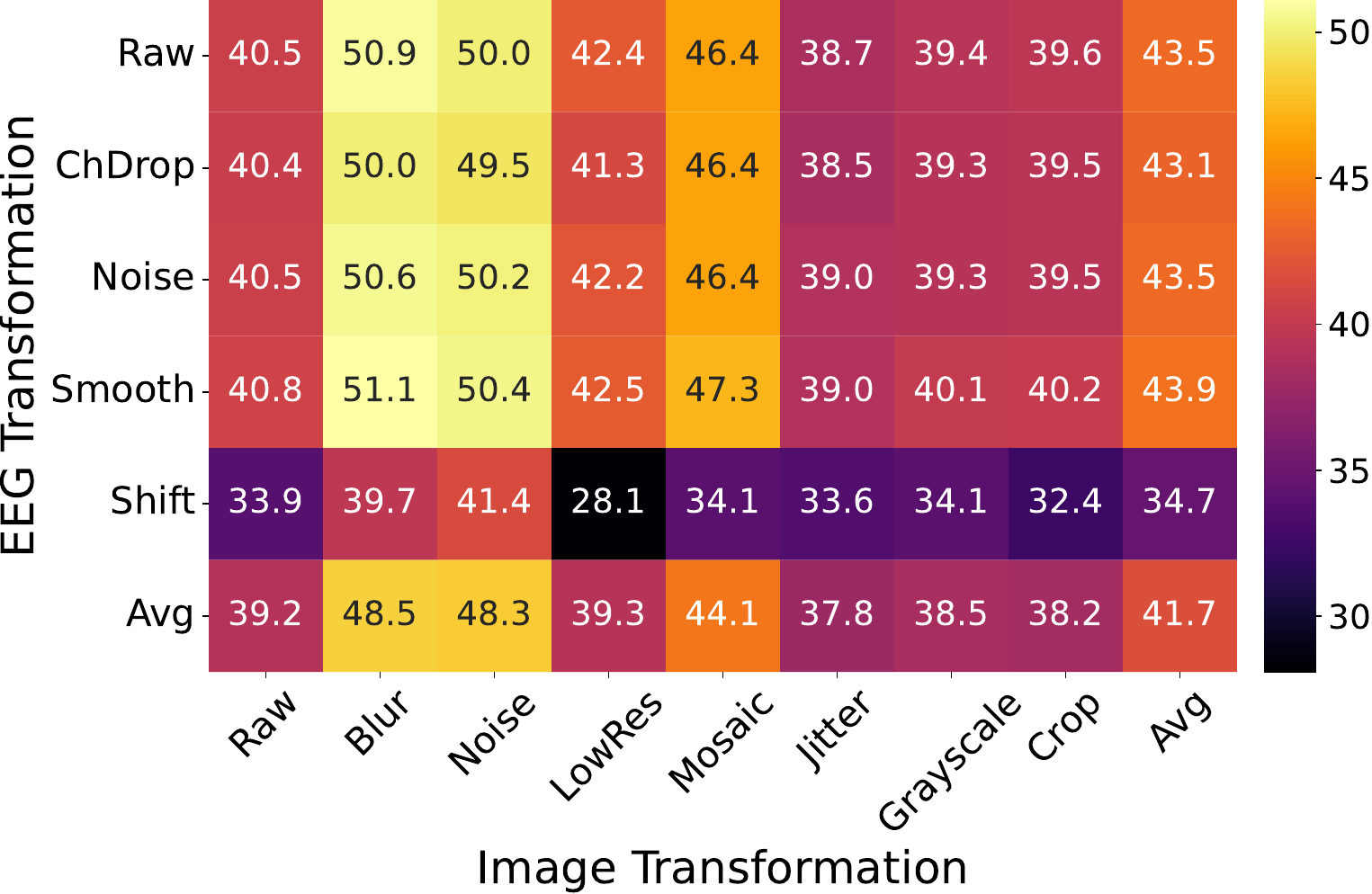}
    \caption{Top-1 accuracy (\%) for different transformation combinations.}
    \label{fig:prior-matrix}
\end{figure}

We evaluated the effectiveness of various image and EEG transformations used in CPA. Figure~\ref{fig:prior-matrix} illustrates the Top-1 accuracy results of different transformation combinations.

\subsubsection{Image Transformation}

\begin{figure}[htbp]
    \centering
    \includegraphics[width=\linewidth]{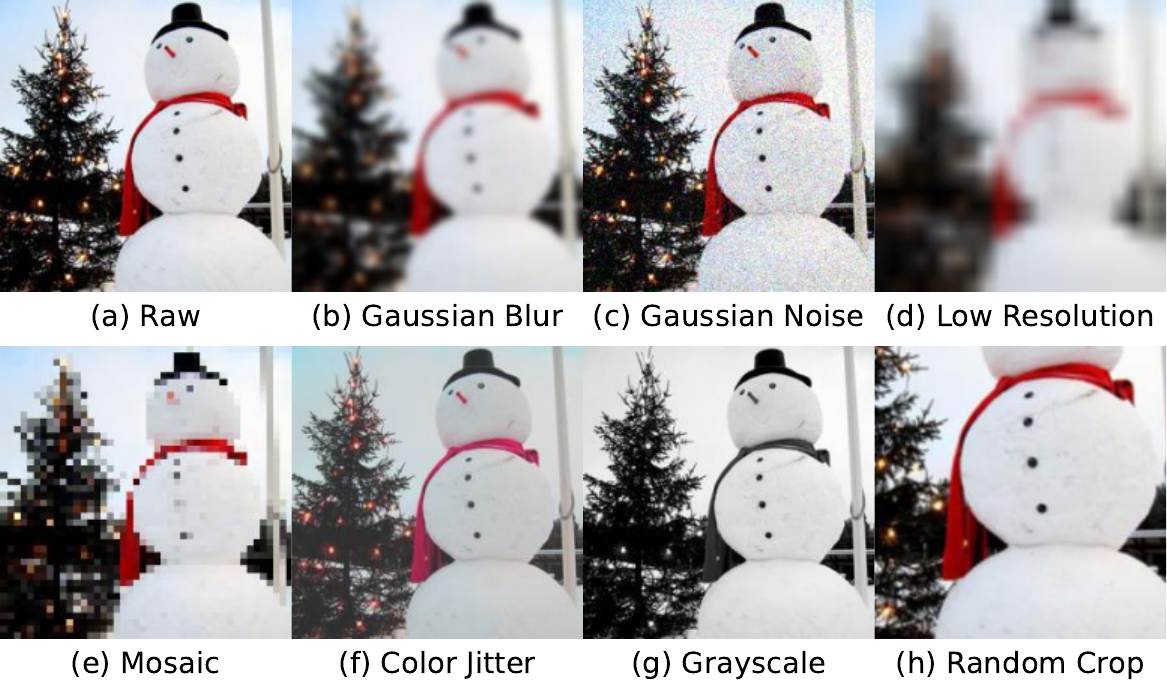}
    \caption{Visual examples of applied image transformations.}
    \label{fig:img-transformation}
\end{figure}

Figure~\ref{fig:img-transformation} displays the visual examples of different image transformation techniques. Gaussian blur, Gaussian noise, low resolution, and mosaic contribute positively to retrieval performance. In contrast, color jitter, grayscale, and random cropping lead to a performance decline. We hypothesize that the former group preserves high-level semantic content while mitigating low-level pixel variations. The negative effects of jitter and grayscale indicate that human perception is sensitive to color information, aligning with findings from current neuroscience research~\cite{Teichmann6779}. Cropping may remove crucial semantic regions, thereby disrupting the image’s structural or semantic integrity.

\subsubsection{EEG Transformation}
EEG transformations provide limited improvement compared to image transformations. Among them, only smoothing consistently enhances performance. Temporal shifting may disturb the temporal dynamics of EEG signals, impairing semantic representation. The beneficial effect of smoothing likely stems from the inherently low SNR of EEG data, as smoothing can mitigate noise interference.

\subsection{Effect of Multi-Transformation Feature Fusion}
We assessed the impact of fusing image features obtained through different transformations. The number of fused transformations ranges from one to seven, following the order shown in Figure~\ref{fig:img-transformation}. As indicated in Table~\ref{table:fusion-comparison}, accuracy improves as the number of fused transformations increases from one to four, then slightly declines from five to seven. This observation is consistent with the earlier results on individual transformation effects. Overall, these findings suggest that fusing diverse image features is beneficial up to a certain point.

\begin{table}[htbp]
    \centering
    \setlength{\tabcolsep}{1.0mm}
    % \small
    \begin{tabular}{lcccccccc}
    \toprule
    \# Transforms & 1 & 2 & 3 & 4 & 5 & 6 & 7 \\
    \midrule
    Top-1 (\%) & 50.9 & 54.2 & 58.4 & \textbf{62.1} & 58.5 & 58.1 & 57.0\\
    Top-5 (\%) & 81.4 & 84.5 & 88.1 & \textbf{89.8} & 88.1 & 87.1 & 86.0\\
    \bottomrule
    \end{tabular}
    \caption{Top-1 and Top-5 accuracy (\%) under different numbers of fused transformations.}
    \label{table:fusion-comparison}
\end{table}

\subsection{Effect of Different Projector Designs}
\begin{figure}[htbp]
    \centering
    \includegraphics[width=0.8\linewidth]{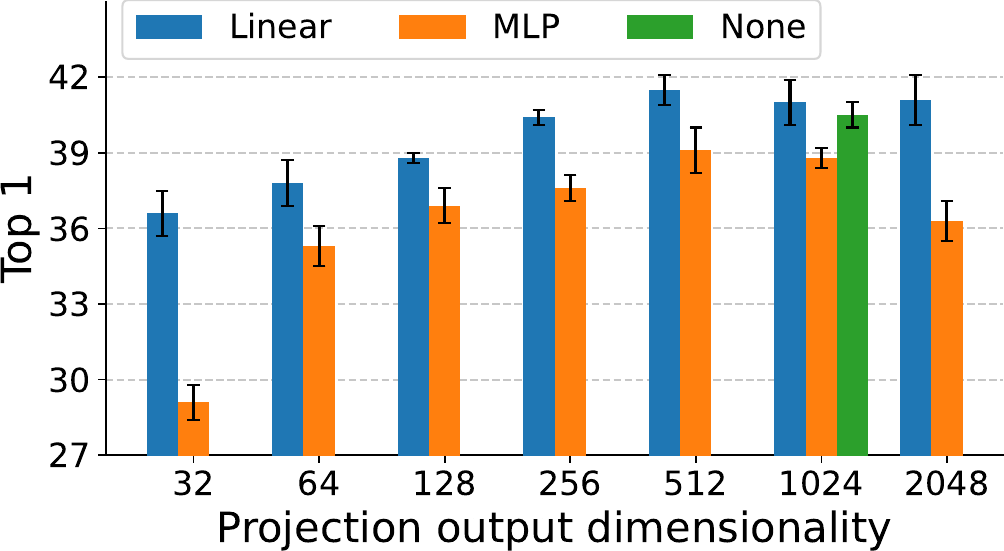}
    \caption{Top-1 accuracy (\%) across different projector types and feature dimensions.}
    \label{fig:projector}
\end{figure}

We evaluated the performance of both linear projection and MLP-based projectors with varying output feature dimensions. As shown in Figure~\ref{fig:projector}, the 512-dimensional linear projection yields the best result. We conjecture this reflects a balance between the model’s expressive capacity and the preservation of information from the pretrained image encoder. Higher-dimensional projectors may lead to overfitting or redundancy, while lower-dimensional ones may be too constrained to capture shared semantics. A larger EEG-image dataset could potentially alleviate this trade-off.

\subsection{Generalization across Encoder Variants}
\begin{figure}[htbp]
    \centering
    \includegraphics[width=\linewidth]{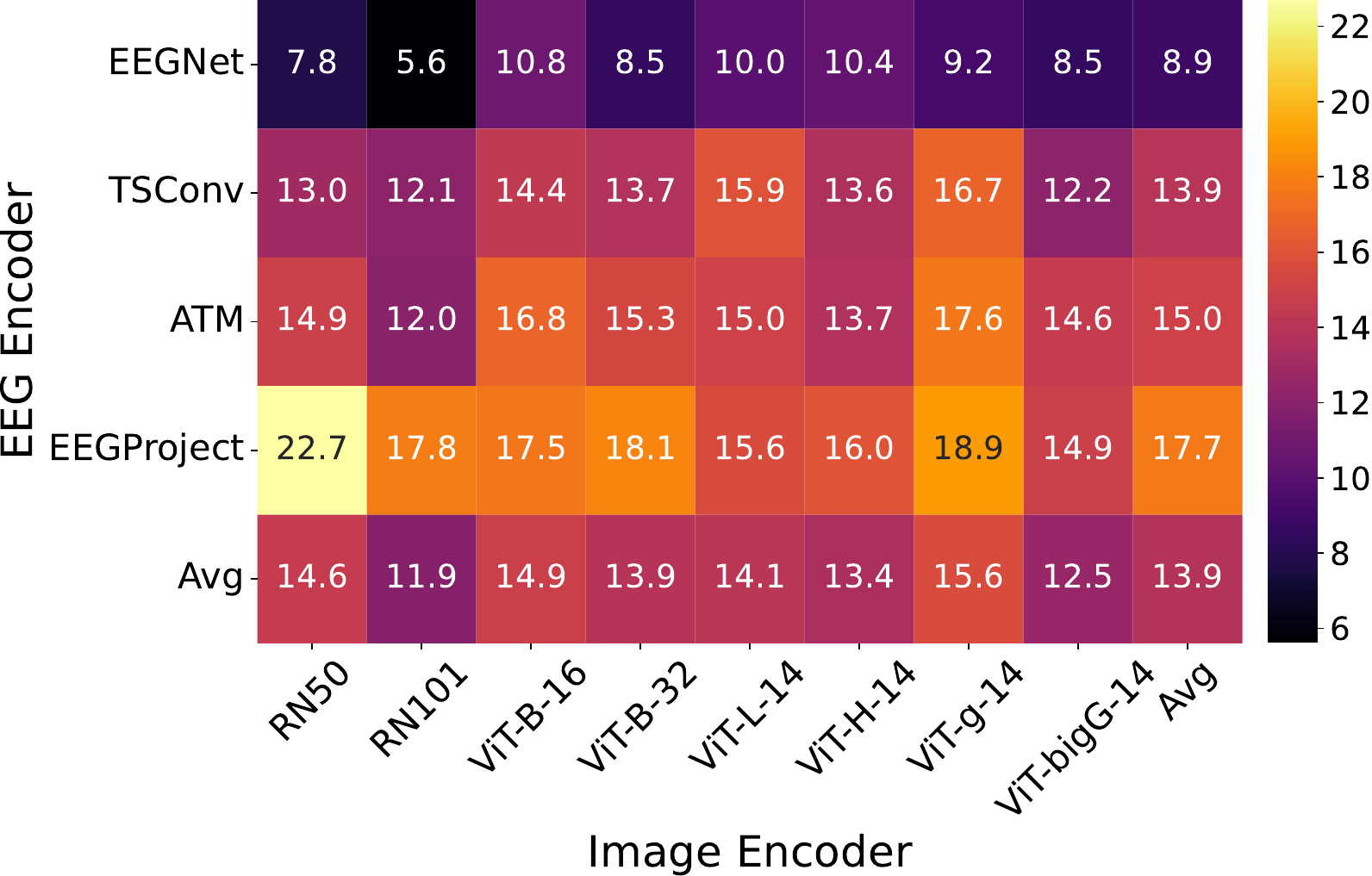}
    \caption{Top-1 accuracy improvement (\%) for various encoder architectures.}
    \label{fig:encoder-matrix}
\end{figure}

To demonstrate the generalizability of our method across different encoder architectures, we implemented several image and EEG encoders. Figure~\ref{fig:encoder-matrix} presents the accuracy improvements achieved with our framework across different encoder combinations.

\subsection{Ablation Study on Core Framework Components}
\begin{table}[ht]
\centering
\begin{tabular}{ccccc}
\toprule
\multicolumn{2}{c}{CPA} & \multirow{2}{*}{SSP} & \multirow{2}{*}{Top-1} & \multirow{2}{*}{Top-5} \\
% \cmidrule(lr){1-2}
Image Prior & EEG Prior & & & \\
\midrule
\xmark & \xmark & \xmark & 40.5 & 72.2 \\
\cmark & \xmark & \xmark & 60.0 & 89.1 \\
\xmark & \cmark & \xmark & 40.8 & 72.7 \\
\xmark & \xmark & \cmark & 41.5 & 73.5 \\
\xmark & \cmark & \cmark & 41.8 & 73.6 \\
\cmark & \xmark & \cmark & 62.1 & 89.8 \\
\cmark & \cmark & \xmark & 60.8 & 89.8 \\
\cmark & \cmark & \cmark & \textbf{63.2} & \textbf{89.9} \\
\bottomrule
\end{tabular}
\caption{Ablation study on the core components of NeuroBridge.}
\label{table:main-components}
\end{table}

To assess the contributions of each component, we performed ablation studies by removing the CPA (including image and EEG priors) and SSP modules. As shown in Table~\ref{table:main-components}, all components positively impact overall performance. The image prior has the greatest effect, likely due to the rich visual information and benefits of pretrained image models.

\subsection{Semantic Alignment Analysis and Visualization}
\begin{figure}[htbp]
    \centering
    \includegraphics[width=\linewidth]{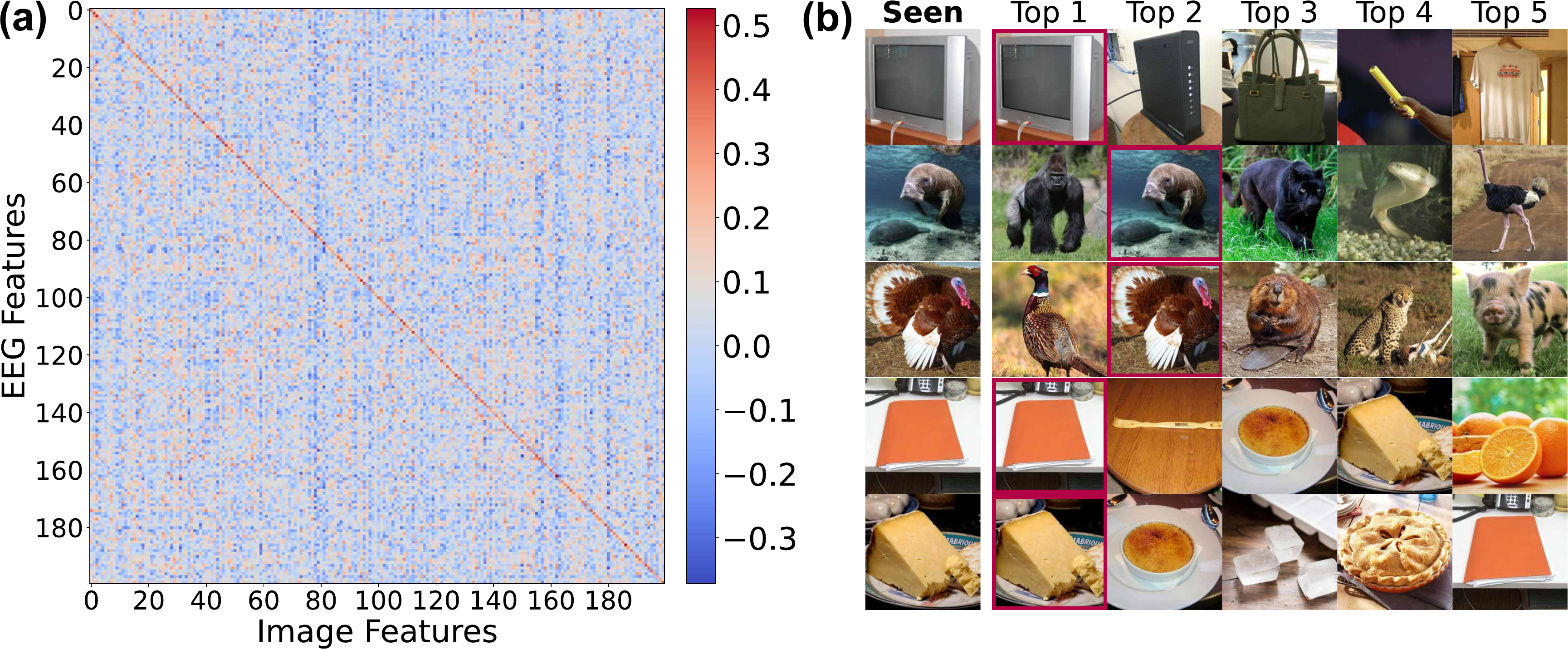}
    \caption{Semantic analysis on subject 8. (a) Similarity matrix. (b) Retrieved samples.}
    \label{fig:semantic-analysis}
\end{figure}

To evaluate the semantic alignment between EEG and visual representations, we computed similarity scores for all 200 test concepts from Subject 8. As illustrated in Figure~\ref{fig:semantic-analysis} (a), the similarity matrix reveals strong alignment across modalities. Furthermore, the retrieval examples in Figure~\ref{fig:semantic-analysis} (b) demonstrate semantic consistency; for example, when the query image depicts an animal, the Top-5 retrieved images are also animals, and a similar trend is observed for food-related images.

\section{Conclusion}
We present NeuroBridge, a unified framework that effectively bridges EEG signals and visual representations for zero-shot neural decoding. By integrating Cognitive Prior Augmentation and Shared Semantic Projector, the method simulates cognitive variability and aligns cross-modal features in a shared semantic space. Experiments demonstrate significant performance gains over existing approaches, confirming the efficacy and generalizability of our framework in both intra-subject and inter-subject settings.
\subsubsection{Limitations}
While NeuroBridge shows strong performance, it has limitations. Manually designed augmentation may insufficiently capture cognitive variability, and reliance on pretrained visual encoders can introduce vision-language biases. These issues suggest future work directions, including adaptive augmentation, data-driven alignment, and scaling up data to improve generalization.

\bigskip

\bibliography{aaai2026}

\clearpage

\appendix

\twocolumn[
\begin{center}

{\LARGE \textbf{Appendix}\\[2pt]
\vspace*{0.3em}
\large Supplemental Material for “NeuroBridge: Bio-Inspired Self-Supervised EEG-to-Image Decoding”\par}
\vspace*{2.0em}
\end{center}
]

\section{A. Experimental Details}

\subsection{A.1. Datasets for experiments}
\subsubsection{THINGS-EEG2}
We conducted our experiments using the THINGS-EEG2 dataset~\cite{gifford2022large}, which includes data from 10 human subjects performing a visual task under the Rapid Serial Visual Presentation (RSVP) paradigm. Each subject participated in four experimental sessions. The training set comprises 1,654 concepts, with each concept represented by 10 unique images and each image repeated four times, resulting in 66,160 trials per subject before averaging. The test set includes 200 concepts, each represented by a single image repeated 80 times, totaling 16,000 trials per subject before averaging. Averaging the repetitions of each image yields 16,540 training samples and 200 test samples per subject.

EEG signals were recorded at a sampling rate of 1000 Hz and filtered to the frequency range of 0.1–100 Hz. The data were segmented into epochs spanning 0 to 1000 ms post-stimulus, with baseline correction performed using the mean signal from the 200 ms pre-stimulus interval. The epoched data were then downsampled to 250 Hz for further analysis. To enhance the signal-to-noise ratio (SNR), all EEG repetitions were averaged in both the training and test sets. The final preprocessed data were stored in float32 format to balance storage efficiency and computational performance.

\subsubsection{THINGS-MEG}
We additionally conducted experiments on the THINGS-MEG dataset~\cite{hebart2023things}, which contains data collected from four human participants performing the same visual recognition task. The main experimental image set spans 1,854 concepts, with 12 distinct images per concept presented once. To construct the zero-shot evaluation setting, we remove all images belonging to the 200 test concepts from this set, leaving 1,654 training concepts and 19,848 training samples per subject. The test set comprises 200 concepts, each represented by a single image repeated 12 times; these repetitions are averaged to obtain 200 test samples per subject.

MEG signals were recorded at 1,200 Hz using a CTF 275 system, with 272 functioning channels. Each trial consisted of a 500 ms stimulus presentation followed by a 1,000 $\pm$ 200 ms inter-stimulus interval. The official preprocessing pipeline~\cite{hebart2023things} applied a 0.1--40 Hz band-pass filter and segmented the data into epochs from -100 to 1,300 ms relative to stimulus onset. Each epoch was baseline normalized by subtracting the mean and dividing by the standard deviation computed over the 100 ms pre-stimulus interval. An additional noisy sensor (MRO11) was excluded, leaving 271 channels, and the epochs were downsampled to 200 Hz.

We use the officially preprocessed epochs and further crop them to the first 1,000 ms after stimulus onset, retaining 200 time points per trial. We then apply channel-wise z-score standardization to both training and test data, using means and standard deviations computed across training trials and time points. To improve SNR, the 12 repetitions of each test image are averaged. The final preprocessed MEG data are stored in float32 format to reduce storage demands and improve I/O efficiency during training and evaluation.

\subsection{A.2. More Implementation Details}
We implement our method using PyTorch and conduct training on two NVIDIA GeForce RTX 3090 GPUs with a batch size of 1,024 for 50 epochs. The optimization is carried out using the AdamW optimizer with a learning rate of 1e-4 and a weight decay of 1e-4. The temperature parameter $\tau$ is set to 0.07, and all reported results represent the average over five independent runs.

For the image encoder, we utilize CLIP with pretrained weights from OpenCLIP, adopting a range of backbones including RN50 (default), RN101, ViT-B-16, ViT-B-32, ViT-L-14, ViT-H-14, ViT-g-14, and ViT-bigG-14. The brain encoder comprises EEGNet, TSConv, ATM, and EEGProject, with EEGProject used by default. Table~\ref{table:eeg-encoders} and Table~\ref{table:img-encoders} present the details of the encoders.

\setcounter{table}{3}
\begin{table}[ht]
\centering
\begin{tabular}{ccccc}
\toprule
EEG Encoder & Channels & Parameters & Emb Dim\\
\midrule
EEGNet & 17    & 2.34M & 1024 \\
EEGNet & 63    & 2.34M & 1024 \\
TSConv & 17    & 2.56M & 1024 \\
TSConv & 63    & 2.63M & 1024 \\
ATM   & 17    & 3.13M & 1024 \\
ATM   & 63    & 3.20M & 1024 \\
EEGProject & 17    & 2.44M & 1024 \\
EEGProject & 63    & 17.18M & 1024 \\
\bottomrule
\end{tabular}
\caption{Details of different EEG encoders.}
\label{table:eeg-encoders}
\end{table}

\begin{table}[ht]
\centering
\begin{tabular}{ccccc}
\toprule
Image Encoder & Parameters & Emb Dim\\
\midrule
RN50  & 38.32M & 1024 \\
RN101 & 56.26M & 512 \\
ViT-B-16 & 86.19M & 512 \\
ViT-B-32 & 87.85M & 512 \\
ViT-L-14 & 303.97M & 768 \\
ViT-H-14 & 632.08M & 1024 \\
ViT-g-14 & 1012.65M & 1024 \\
ViT-bigG-14 & 1844.91M & 1280 \\
\bottomrule
\end{tabular}
\caption{Details of different Image encoders.}
\label{table:img-encoders}
\end{table}

For the THINGS-EEG2 dataset, considering the differences in data quantity and model complexity between inter-subject and intra-subject settings, we adopt specific configurations for each. In the intra-subject setting, we use a learning rate of 1e-4 and restrict the EEG input to 17 electrodes located over the parietal and occipital regions, specifically: P7, P5, P3, P1, Pz, P2, P4, P6, P8, PO7, PO3, POz, PO4, PO8, O1, Oz, and O2. The EEG encoder in this case is EEGProject. In contrast, the inter-subject setting uses a learning rate of 1e-4 with all 63 EEG channels as input, and the EEG encoder is set to TSConv.

For the THINGS-MEG dataset, the number of recording channels is substantially larger than in THINGS-EEG2. To control the overall model parameter scale, we employ TSConv as the neural encoder and utilize all 271 sensors in both intra-subject and inter-subject experimental settings. All other training configurations remain consistent with those used for THINGS-EEG2.

\begin{table*}[t]
\centering
\small
\begin{tabular}{lc|cccc|c}
\toprule
\multicolumn{2}{l|}{Method} & Sub 1 & Sub 2 & Sub 3 & Sub 4 & Average \\
\midrule
\multicolumn{7}{c}{\textbf{Intra-Subject: train and test on one subject}} \\
\midrule
\multirow{2}{*}{NICE} & Top-1 & 9.6 & 18.5 & 14.2 & 9.0 & 12.8\\
& Top-5 & 27.8 & 47.8 & 41.6 & 26.6 & 36.0\\
\midrule
\multirow{2}{*}{UBP} & Top-1 & 15.0 & 46.0 & 27.3 & \textbf{18.5} & 26.7\\
& Top-5 & 38.0 & 80.5 & 59.0 & \textbf{43.5} & 55.2\\
\midrule
\multirow{2}{*}{\textbf{NeuroBridge (Ours)}} & Top-1 & \textbf{16.5} & \textbf{53.7} & \textbf{40.4} & 18.1 & \textbf{32.2}\\
& Top-5 & \textbf{41.6} & \textbf{85.3} & \textbf{73.2} & 43.1 & \textbf{60.8}\\
\midrule
\multicolumn{7}{c}{\textbf{Inter-Subject: leave one subject out for test}} \\
\midrule
\multirow{2}{*}{UBP} & Top-1 & 2.0 & 1.5 & 2.7 & \textbf{2.5} & 2.2\\
& Top-5 & 5.7 & \textbf{17.2} & 10.5 & 8.0 & 10.4\\
\midrule
\multirow{2}{*}{\textbf{NeuroBridge (Ours)}} & Top-1 & \textbf{4.3} & \textbf{3.6} & \textbf{3.0} & \textbf{2.5} & \textbf{3.4}\\
& Top-5 & \textbf{13.1} & 15.6 & \textbf{11.2} & \textbf{11.3} & \textbf{12.8}\\
\bottomrule
\end{tabular}
\caption{Overall accuracy (\%) of 200-way zero-shot retrieval on THINGS-MEG: Top-1 and Top-5.}
\label{table:meg}
\end{table*}

For Cognitive Prior Augmentation (CPA), we employ various image augmentations including Gaussian blur, Gaussian noise, low resolution, mosaic, color jitter, grayscale, and random crop. EEG augmentations consist of channel dropout, noise addition, smoothing, and temporal shifting. By default, Gaussian blur, Gaussian noise, low resolution, and mosaic are applied to images, and smoothing is applied to EEG signals.

In the Shared Semantic Projector (SSP) module, we experiment with both linear projections and multi-layer perceptrons (MLPs) with varying feature dimensions.

For the ablation studies, Figure~\ref{fig:prior-matrix} evaluates different combinations of image and EEG transformations. Table~\ref{table:fusion-comparison} evaluates multi-transformation image feature fusion using Image Prior and SSP, with EEG Prior disabled. Figure~\ref{fig:projector} focuses solely on SSP projector designs. All ablation experiments are conducted under the intra-subject setting.

\section{B. Additional Results and Details}

To assess the generalization capacity of NeuroBridge, we further evaluate it on the THINGS-MEG dataset. Our method maintains consistent state-of-the-art performance under this setting, and the results are summarized in Table~\ref{table:meg}.

Table~\ref{table:loss} reports the performance under different $\ell_2$ normalization strategies, clearly indicating that the asymmetric loss formulation leads to performance improvements.

\begin{table}[ht]
\centering
\begin{tabular}{ccccc}
\toprule
Method & $\ell_2$-EEG & $\ell_2$-Image & Top-1 & Top-5 \\
\midrule
Plain & \xmark & \xmark & 54.4 & 85.5 \\
Sym & \cmark & \cmark & 46.4 & 79.0 \\
Inv-Asym & \cmark & \xmark & 38.6 & 71.7 \\
\textbf{Asym} & \xmark & \cmark & \textbf{63.2} & \textbf{89.9} \\
\bottomrule
\end{tabular}
\caption{Comparison of $\ell_2$ normalization strategies on EEG and image representations.}
\label{table:loss}
\end{table}

The detailed results of Top-1 and Top-5 accuracy for both Vanilla and NeuroBridge are shown in Figures~\ref{fig:top1-vanilla}--\ref{fig:top5-improvement}.

\setcounter{figure}{7}
\begin{figure}[htbp]
\centering
\includegraphics[width=\linewidth]{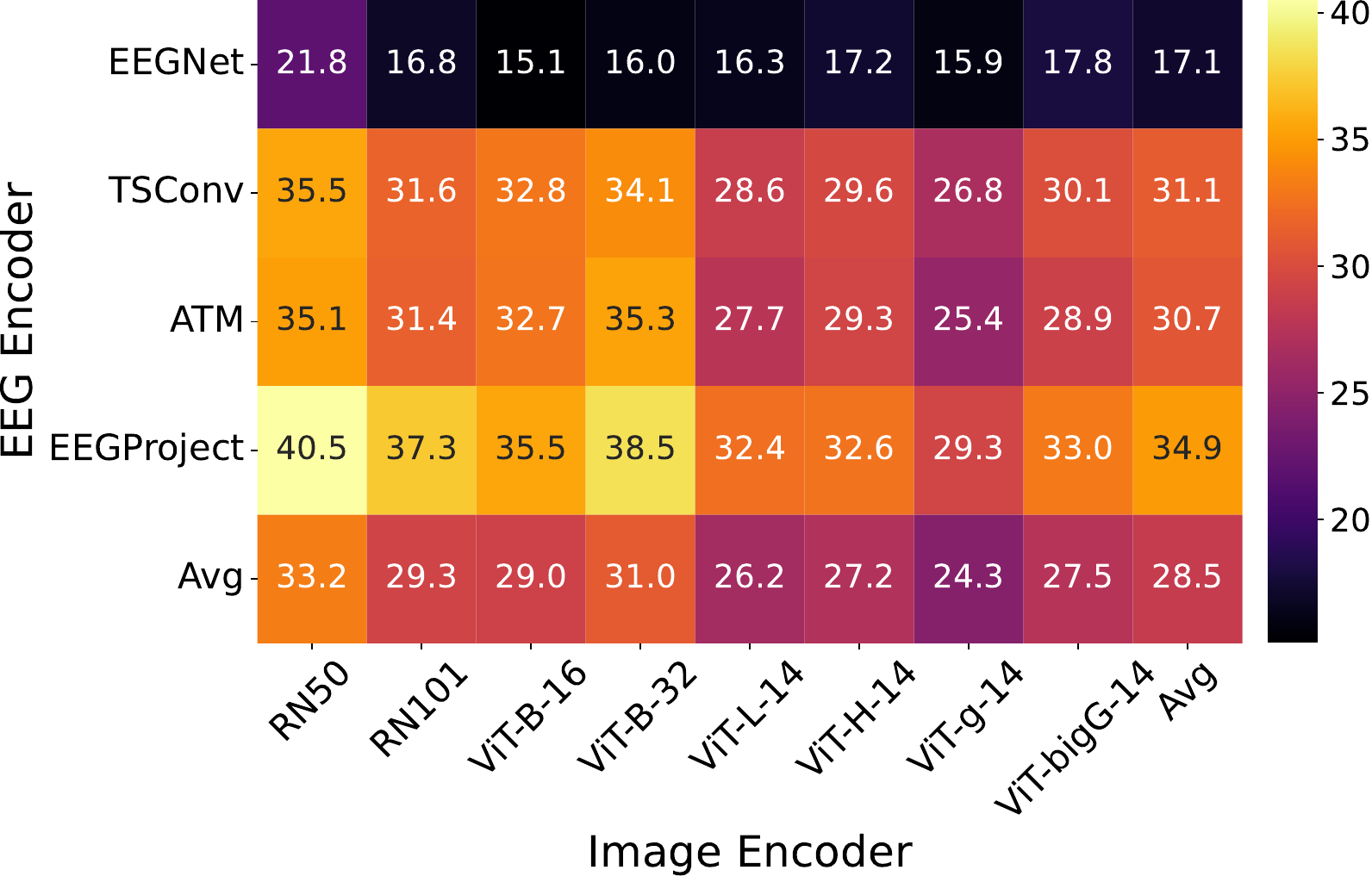}
\caption{Top-1 accuracy (\%) of Vanilla on the THINGS-EEG2 dataset.}
\label{fig:top1-vanilla}
\end{figure}

\begin{figure}[htbp]
\centering
\includegraphics[width=\linewidth]{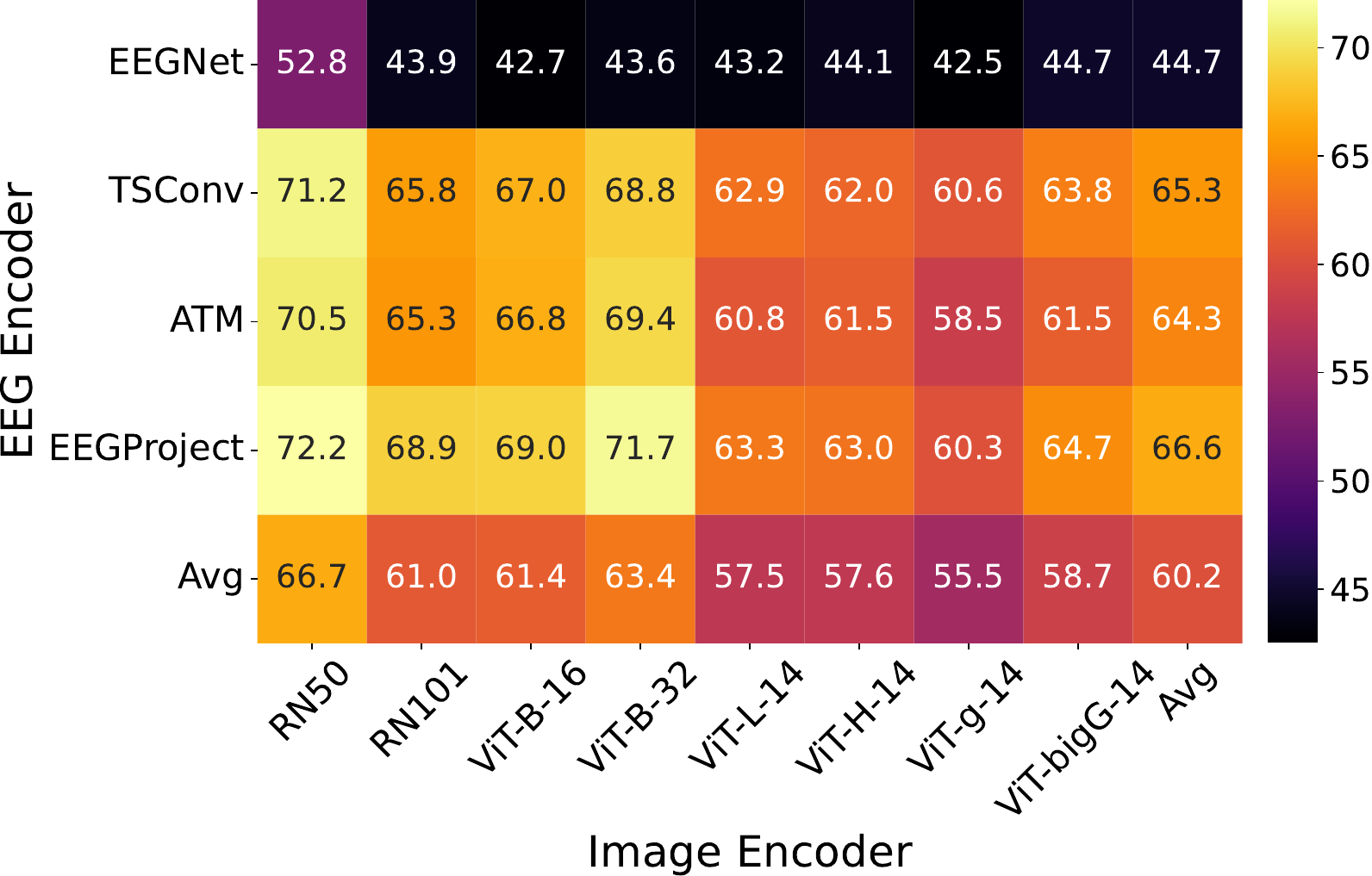}
\caption{Top-5 accuracy (\%) of Vanilla on the THINGS-EEG2 dataset.}
\label{fig:top5-vanilla}
\end{figure}

Given that contrastive learning is highly sensitive to batch size and temperature, we evaluated the performance of NeuroBridge under varying configurations of these two factors, as presented in Table~\ref{table:batch-comparison} and Table~\ref{table:temperature-comparison}. Compared to standard contrastive learning, NeuroBridge exhibits greater robustness to such variations. This resilience may be attributed to the use of a frozen, pre-trained image encoder, which mitigates the impact of hyperparameter fluctuations.

\begin{figure}[htbp]
\centering
\includegraphics[width=\linewidth]{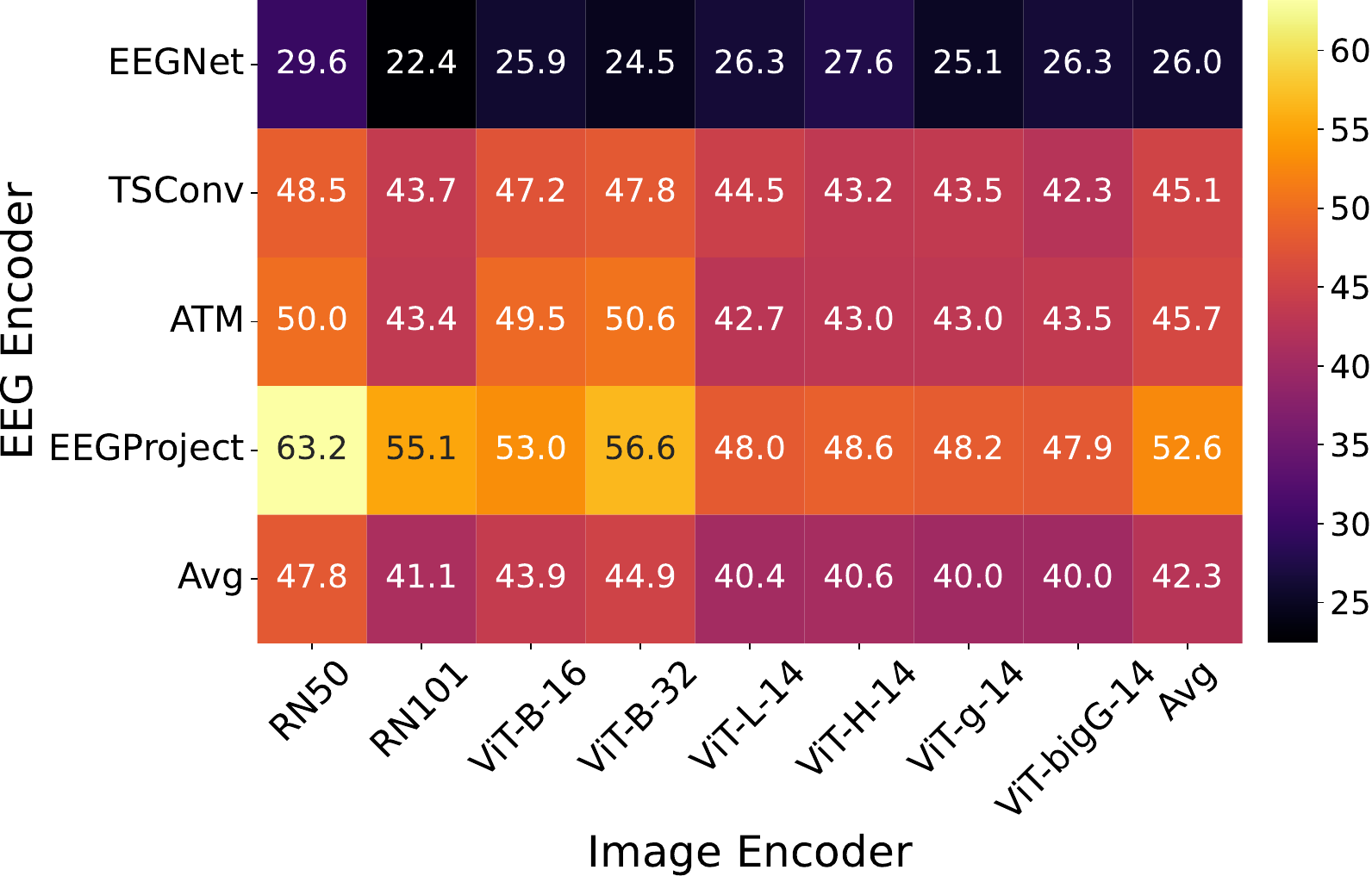}
\caption{Top-1 accuracy (\%) of NeuroBridge on the THINGS-EEG2 dataset.}
\label{fig:top1-neurobridge}
\end{figure}

\begin{figure}[htbp]
\centering
\includegraphics[width=\linewidth]{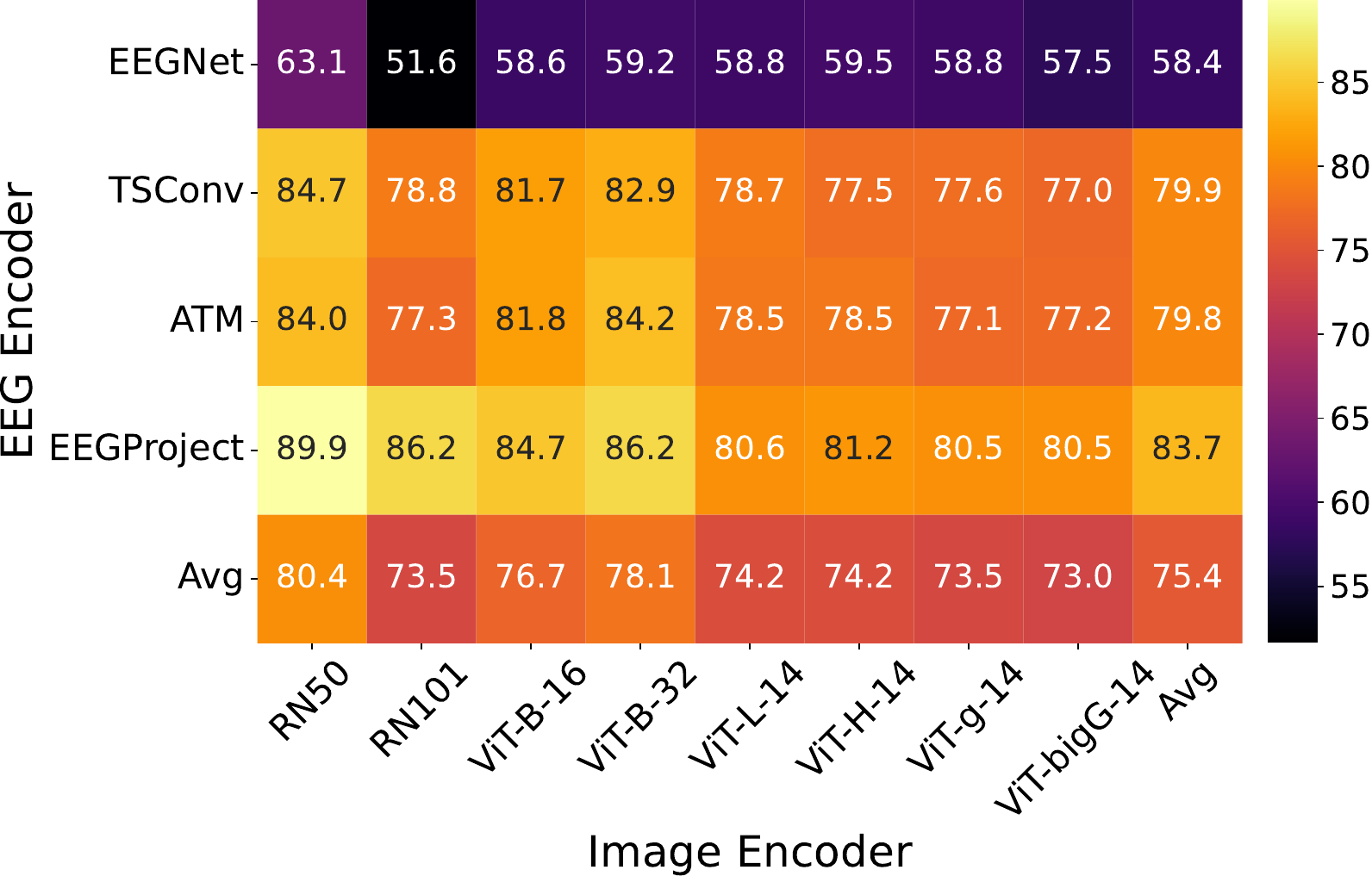}
\caption{Top-5 accuracy (\%) of NeuroBridge on the THINGS-EEG2 dataset.}
\label{fig:top5-neurobridge}
\end{figure}

\begin{figure}[htbp]
\centering
\includegraphics[width=\linewidth]{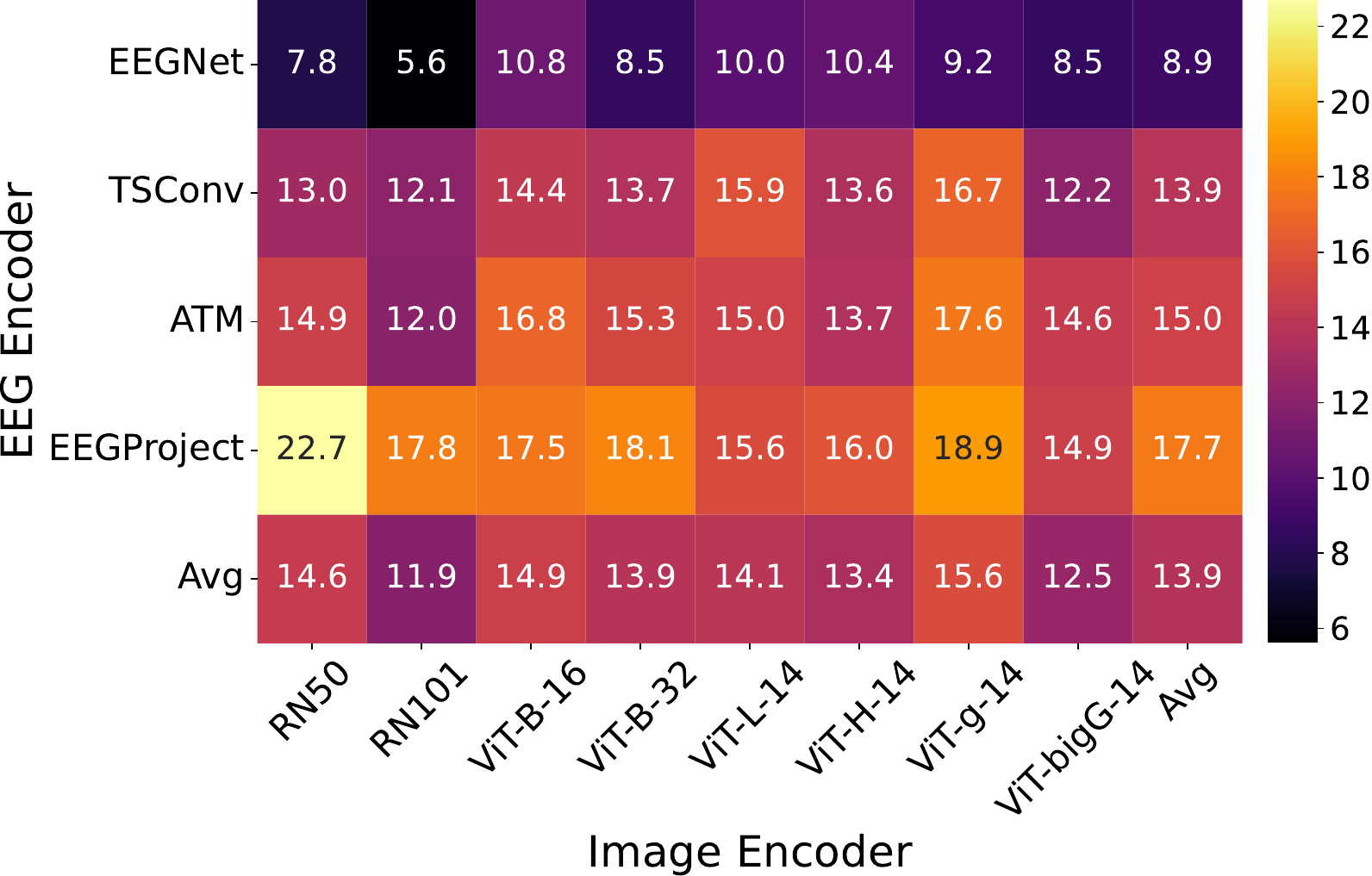}
\caption{Top-1 accuracy improvement (\%) on the THINGS-EEG2 dataset.}
\label{fig:top1-improvement}
\end{figure}

\begin{figure}[htbp]
\centering
\includegraphics[width=\linewidth]{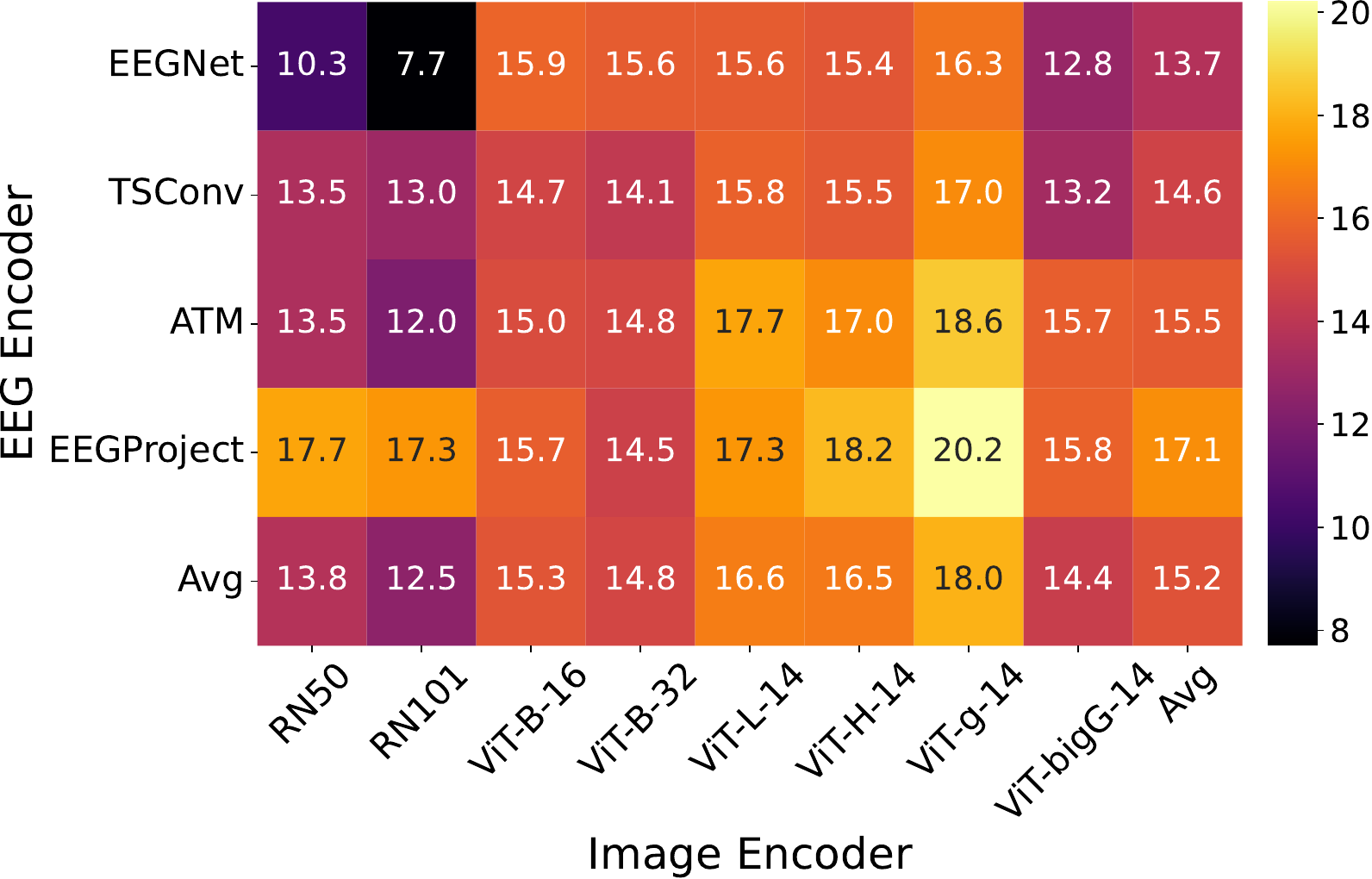}
\caption{Top-5 accuracy improvement (\%) on the THINGS-EEG2 dataset.}
\label{fig:top5-improvement}
\end{figure}

\begin{table}[htbp]
    \centering
    \setlength{\tabcolsep}{1.0mm}
    % \small
    \begin{tabular}{lcccccccc}
    \toprule
    Batch Size & 32 & 64 & 128 & 256 & 512 & 1024 & 2048\\
    \midrule
    Top-1 (\%) & 54.6 & 54.9 & 57.6 & 59.4 & 60.8 & \textbf{63.2} & 62.2\\
    Top-5 (\%) & 83.9 & 84.8 & 86.5 & 87.8 & 89.1 & \textbf{89.9} & 89.8\\
    \bottomrule
    \end{tabular}
    \caption{Top-1 and Top-5 accuracy (\%) under different batch sizes.}
    \label{table:batch-comparison}
\end{table}

\begin{table*}[tp]
    \centering
    \setlength{\tabcolsep}{1.0mm}
    \small
    \begin{tabular}{lcccccccc}
    \toprule
    Temperature & Learnable & 0.001 & 0.005 & 0.01 & 0.05 & 0.1 & 0.5\\
    \midrule
    Top-1 (\%) & 62.1 & 59.5 & 61.0 & 60.7 & 62.0 & 62.2 & \textbf{63.6}\\
    Top-5 (\%) & 90.2 & 88.9 & 89.0 & 89.2 & 89.8 & 89.6 & \textbf{91.0}\\
    \bottomrule
    \end{tabular}
    \caption{Top-1 and Top-5 accuracy (\%) under different temperatures.}
    \label{table:temperature-comparison}
\end{table*}

\begin{figure*}[tp]
    \centering
    \includegraphics[width=\textwidth]{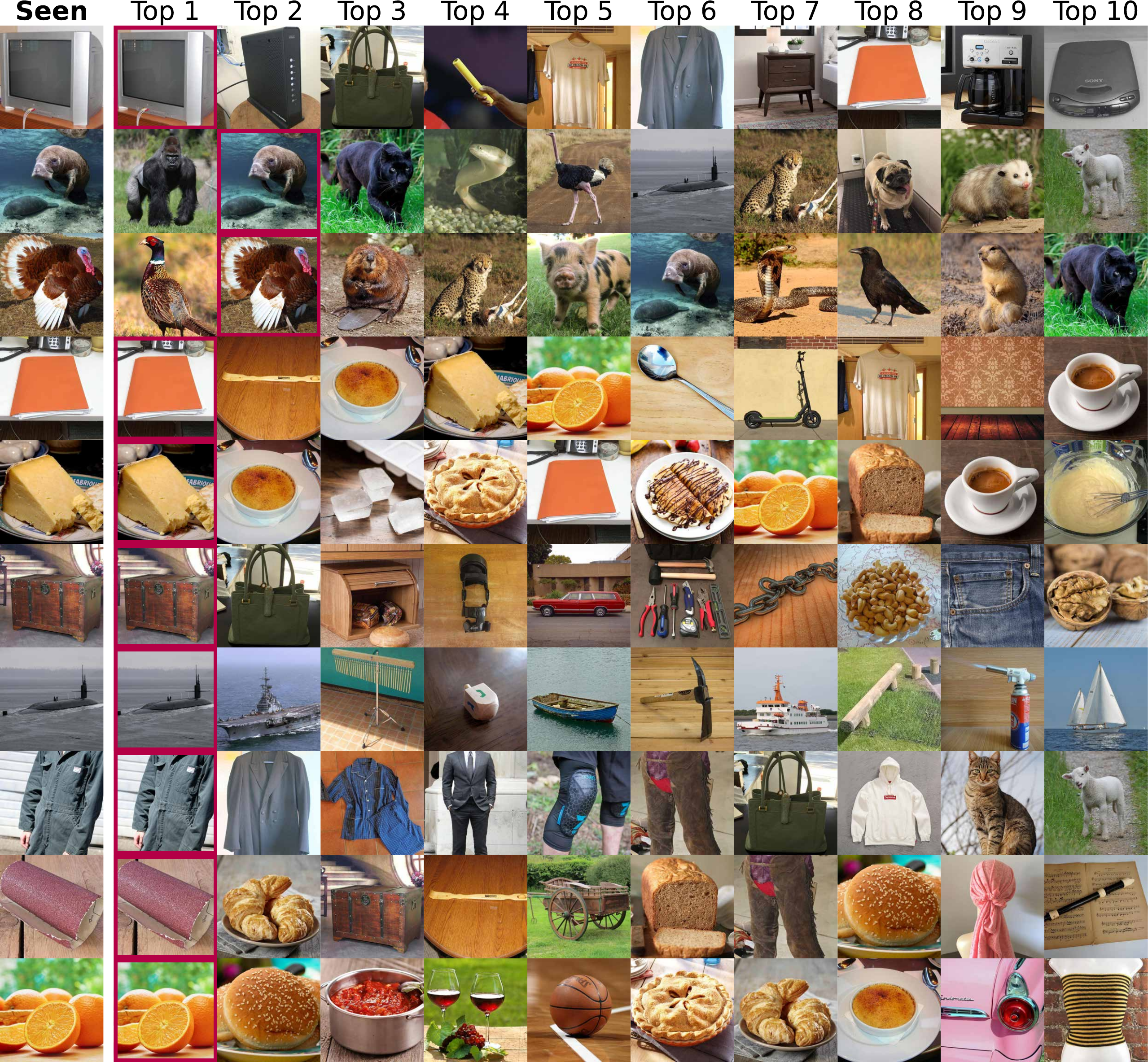}
    \caption{More retrieved samples.}
    \label{fig:retrieved-samples}
\end{figure*}

Figure~\ref{fig:retrieved-samples} presents further examples of retrieval results to illustrate the model's performance.

\end{document}